\documentclass{article}
\usepackage{float}

\usepackage{microtype}
\usepackage{graphicx}
\usepackage{subcaption}
\usepackage{booktabs} 
\usepackage{xurl}

\usepackage{capt-of}

\usepackage{hyperref}

\usepackage[preprint]{icml2026}

\usepackage{amsmath}
\usepackage{amssymb}
\usepackage{mathtools}
\usepackage{amsthm}
\usepackage{afterpage}
\usepackage{stfloats}

\usepackage[capitalize,noabbrev]{cleveref}

\theoremstyle{plain}

\theoremstyle{definition}

\theoremstyle{remark}

\icmltitlerunning{Belief or Circuitry?}

\begin{document}

\twocolumn[
  \icmltitle{Belief or Circuitry? Causal Evidence for In-Context Graph Learning}
  \icmlsetsymbol{equal}{*}

  \begin{icmlauthorlist}
    \icmlauthor{Katharine Kowalyshyn}{equal,yyy}
    \icmlauthor{Timothy Duggan}{equal,yyy}
    \icmlauthor{Daniel Little}{yyy}
    \icmlauthor{Michael C. Hughes}{yyy}
  \end{icmlauthorlist}

  \icmlaffiliation{yyy}{Department of Computer Science, Tufts University, Medford, United States}

  \icmlcorrespondingauthor{Katharine Kowalyshyn}{katharine.kowalyshyn@tufts.edu}

  \icmlkeywords{Mechanistic Interpretability, In-Context Learning}

  \vskip 0.3in
]



\printAffiliationsAndNotice{}  

\begin{abstract}
How do LLMs learn in-context? Is it by pattern-matching recent tokens, or by 
inferring latent structure? We probe this question using a toy graph random-walk across two competing graph structures. This task's answer is, in principle, decidable: either the model tracks 
global topology, or it copies local transitions. We present two lines of 
evidence that neither account alone is sufficient. 
First, reconstructing the internal representation structure via PCA reveals that at intermediate mixture ratios, both graph topologies are encoded in orthogonal principal subspaces simultaneously. This pattern is difficult to reconcile with purely local transition copying. Second, residual-stream activation patching and graph-difference steering causally intervene on this graph-family signal: late-layer patching almost fully transfers the clean graph preference, while linear steering moves predictions in the intended direction and fails under norm-matched and label-shuffled controls. Taken together, our findings are most consistent with a dual-mechanism account in which genuine structure inference and induction circuits operate in parallel. Code is available at \href{https://anonymous.4open.science/r/do-llms-infer-graphs-C67A}{this link}.
\end{abstract}

\section{Introduction \& Related Works}
Since the advent of LLMs, in-context learning (ICL) has remained an area of research that stumps the community. Over the past five years, numerous investigations into ICL have yielded interesting results: alignment \cite{anwar2024foundationalchallengesassuringalignment, lin2023unlockingspellbasellms}, jailbreaking \cite{polyakov2026involuntaryincontextlearningexploiting}, and demonstration 
selection \cite{qin2024incontextlearningiterativedemonstration} are just a few areas that have prioritized it as a subfield of study \cite{dong-etal-2024-survey}. 

Within the subfield of mechanistic interpretability (MechInterp), there is 
yet to be a concrete consensus on this topic. Originally posed as a debate 
between inference over latent structure or shallow pattern-matching circuits 
by~\citet{olsson2022incontextlearninginductionheads}, ICL has seen numerous 
recent works in MechInterp from theoretical accounts framing ICL as 
implicit Bayesian inference over latent concepts~\citep{xie2022explanationincontextlearningimplicit}, 
to mechanistic studies of induction head formation and 
diversity~\citep{singh2024needsrightinductionhead}, to causal evidence that ICL decomposes 
into separable task schema and input-output binding 
mechanisms~\citep{kim2025taskschemabindingdouble}.

Recent work from \citet{park2025iclrincontextlearningrepresentations} provided striking evidence for the former: given random-walk traces on an unknown graph as a toy task, Llama-3.1-8B exhibits a \textit{sharp phase transition} in neighbor-prediction accuracy. This transition results in
simultaneously reorganizing its residual-stream geometry to mirror the graph's adjacency structure. If the phase transition reflects latent-structure inference, then LLMs maintain implicit probabilistic world models that can be probed mechanistically. If it reflects induction circuits, the same behavioral signature requires only local copying heads and carries no implications about global representations.

\textbf{Contributions.} Our contributions are a first step to triangulate the answer to this ongoing debate. \textbf{(1)} We replace the flat log-prior of \citet{bigelow2025beliefdynamicsrevealdual} with a complexity-weighted structure-specific prior, recovering a quantitative signature of topology-sensitive structural bias. \textbf{(2)} We expose Llama-3.1-8B to interleaved walks from two competing graphs and show that the belief account's prediction of topology-biased evidence accumulation holds, against the induction circuit account's prediction of symmetric behavior. \textbf{(3)} PCA of residual-stream activations reveals both graph topologies are simultaneously recoverable in orthogonal subspaces at intermediate mixture ratios. \textbf{(4)} Activation patching and steering causally link this representational structure to next-token predictions.


Ultimately, our conclusion is that the question of whether LLMs \textit{believe} or \textit{copy} may be a false dichotomy, and the architecture of that coexistence is what mechanistic interpretability must now explain.

\section{Background}

Recently published works in ICL have taken two simultaneous approaches to investigating this problem. \citet{park2025iclrincontextlearningrepresentations} used a toy task of random walks over a sixteen word grid where nodes are non-semantically related words and edges are connections between word pairs adjacent in the grid. That paper showed that Llama-3.1-8B undergoes a sharp phase transition in neighbor-hit accuracy as context length grows, and that layer-level PCA of node-token representations progressively recovers the true graph topology. This was interpreted as evidence for an implicit Bayesian world model over graph structure.
Following the publication of \citet{park2025iclrincontextlearningrepresentations}, 
two blog posts entered the debate. \citet{arditi2026induction} identified 
specific attention heads in Llama-3.1-8B that implement induction, arguing 
that the phase transition in graph ICL is fully explained by these heads 
accumulating local transition statistics, with no need to posit global 
structure inference. \citet{ransome_2026} replicated these findings and 
extended the analysis to additional graph topologies. We note that this 
second post appeared concurrently with our own work; we refer readers to 
both for complementary perspectives on the mechanistic debate.

Alongside the ongoing debate around work by  \citet{park2025iclrincontextlearningrepresentations}, there was a theoretical Bayesian dynamics approach to ICL published by \citet{bigelow2025beliefdynamicsrevealdual}. These authors fit a sigmoid-shaped parametric function over log-odds evidence accumulation to LLM accuracy curves, treating the model as maintaining a latent binary belief over two hypotheses about the data source. They found evidence for a dual-mechanism account in which both Bayesian updating and induction circuits contribute. 


\section{Behavioral Model}

The central question is whether the LLM's behavior during ICL looks more like 
a Bayesian reasoner accumulating evidence about latent structure, or a 
pattern-matcher copying recent tokens. To probe this, we fit a belief-dynamics model to observed accuracy curves and ask: do the recovered 
parameters tell a story consistent with genuine structural inference?

\subsection{General Framework}

Consider an LLM presented with a context generated by one of $K$ competing 
hypotheses $\mathcal{H} = \{H_1, \ldots, H_K\}$ about the latent 
data-generating structure. We model the LLM as maintaining a latent belief over 
which hypothesis is active, updating that belief as context accumulates. For 
each hypothesis $H_k$, the model's prior skepticism is governed by a log-odds 
term $b_k \in \mathbb{R}$, with $b_k < 0$ encoding initial skepticism. 
Evidence in favor of $H_k$ accumulates sub-linearly with context length $N$, 
giving a predicted accuracy at context length $N$:
\begin{equation}
  \hat{p}_k(N) = p_{0,k} + (q_k - p_{0,k})\,\sigma\!\left(b_k + \gamma_k 
  N^{1-\alpha_k}\right)
  \label{eq:sigmoid-general},
\end{equation}
where $p_{0,k}$ is the pre-transition accuracy under $H_k$, $q_k$ is the 
post-transition accuracy, $\gamma_k > 0$ controls evidence strength, and 
$\alpha_k \in (0,1)$ captures diminishing returns from correlated observations. 
The inflection point $N^*_k = (-b_k/\gamma_k)^{1/(1-\alpha_k)}$ marks the 
context length at which the LLM tips from skepticism to commitment under $H_k$.

A key question is what determines $b_k$, the LLM's initial bias toward or 
against each hypothesis. We propose that $b_k$ is governed by the 
\textit{complexity} of $H_k$: a more complex hypothesis requires more evidence to overcome the prior. Concretely, we parameterize
\begin{equation}
  b_k = b_0 - \lambda \cdot C(H_k),
  \label{eq:prior-general}
\end{equation}
where $b_0$ is a shared baseline log-odds, $\lambda \geq 0$ is a learned 
penalty weight, and $C(H_k)$ is an MDL-inspired complexity measure of hypothesis $H_k$. If $\hat\lambda > 0$, the LLM implicitly penalizes more complex hypotheses, a signature of complexity-sensitive structural inference that topology-agnostic pattern matching cannot produce.

\subsection{Instantiation: Competing Graph Structures}

We instantiate this framework using the graph random-walk task of 
\citet{park2025iclrincontextlearningrepresentations}. The LLM is presented with token sequences generated by random walks over an unknown graph $G$, and must predict the next node, a valid neighbor of the current node. The two competing hypotheses are $H_{\text{grid}}$ and $H_{\text{ring}}$: a $4{\times}4$ grid (16 nodes, 24 edges, degree 2--4) and a 16-node ring (16 edges, uniform degree 2), each with nodes assigned distinct single-token English nouns. The observable accuracy $\hat{p}_k(N)$ is the neighbor-hit probability, the probability that the model's next-token prediction is a valid graph neighbor of the current node under hypothesis $H_k$.

The MDL complexity of each graph hypothesis is naturally given by the length of 
its edge-list encoding:
\begin{equation}
  C(G) = |E(G)| \cdot \lceil \log_2 |V| \rceil \;\text{bits},
\end{equation}
yielding $C(\text{grid}) = 96$ bits and $C(\text{ring}) = 64$ bits. The grid 
costs more to describe because it has more edges. If $\hat\lambda > 0$ with 
$\hat{b}_{\text{grid}} < \hat{b}_{\text{ring}}$, the LLM requires 
disproportionately more context to commit to the denser topology. This is inconsistent 
with induction heads, which accumulate transition statistics uniformly 
regardless of graph structure and predict $\hat\lambda \approx 0$.

To create a genuine competition between hypotheses, we interleave walks from 
both graphs at a controlled mixture ratio $\rho \in [0, 1]$, where $\rho$ is 
the probability that any given 100-token segment is drawn from the ring walk. 
The effective context length for graph $k$ is $\rho_k \cdot N$ (a mean-field 
approximation that holds in expectation across walk realizations) giving
\begin{equation}
  \hat{p}_k(\rho, N) = p_{0,k} + (q_k - p_{0,k})\,
  \sigma\!\bigl(b_k + \gamma_k(\rho_k N)^{1-\alpha_k}\bigr).
\end{equation}
We compare a \textbf{per-graph} parameterization (8 free parameters: $b_0, 
\lambda, \gamma_k, \alpha_k, q_k$ per graph) against a \textbf{mixture-bias} 
ablation (5 parameters) that shares a single sigmoid but linearly interpolates 
the prior: $b(\rho) = (1-\rho)b_{\text{grid}} + \rho\,b_{\text{ring}}$. The 
mixture-bias version can capture prior asymmetry but not topology-specific 
evidence rates, making it a direct test of whether per-graph dynamics are 
needed beyond the prior alone. Model selection uses AIC and BIC; see 
Appendix~\ref{app:upgrade} for estimation details. We use Llama-3.1-8B 
(non-instruct) loaded via TransformerLens~\citep{nanda2022transformerlens}, 
with layer-26 residual-stream activations for representational analyses 
following \citet{park2025iclrincontextlearningrepresentations}.

\section{Experiment 1}

\subsection{Which Sigmoid Fits Best?}

For each (condition, $\rho$) cell, we fit both the baseline and our model with the complexity-weighted prior
to the training walks and evaluate on held-out sequences. 


\subsection{Behavioral Sigmoid Fits}

The grid's inflection point $N^*$ shifts monotonically later as ring evidence increases, precisely the graph-level competition effect the belief account predicts. A flat induction account cannot explain this, since copy heads
accumulate transitions without any topology-aware interaction. The recovered parameters satisfy $\hat\lambda > 0$ and $\hat{b}_{\text{grid}} < \hat{b}_{\text{ring}}$ in both vocabulary conditions, the per-graph parameterization decisively outperforms the mixture-bias ablation on AIC and BIC, and at $\rho = 1$ the ring converges faster than the grid, which is consistent with its lower MDL complexity requiring less evidence to overcome.

\section{Experiment 2}

\subsection{Does the Residual Stream Encode Latent Graph Structure?}

To test whether the behavioral signatures reflect genuine changes in internal
representations, we probe activations directly,
following the representational analysis of
\citet{park2025iclrincontextlearningrepresentations}. For each node $v$, we average activations over all positions where $w_t = v$ at context length $T$,
yielding a class-mean matrix. Projecting these class-mean vectors into PCA space lets us ask whether the low-dimensional geometry recovers the true graph topology, and, critically, whether both graph topologies are simultaneously recoverable at intermediate $\rho$.

To quantify structural alignment we report degree-normalized Dirichlet energy
under the true graph Laplacian $L = D - A$, where $A$ is the adjacency
matrix and $D$ is the diagonal degree matrix. We define $H_T \in \mathbb{R}^{|V| \times d}$ as the matrix of class-mean
activations with rows $\mu_v(T)$ for each node $v$ present in a trailing
context window at length $T$, and $\bar{H}_T$ is the degree-weighted mean.
\begin{equation}
  \mathcal{E}(T) = \mathrm{Tr}(H_T^\top L H_T)
  = \frac{1}{2}\sum_{i,j} A_{ij}\lVert \mu_i(T) - \mu_j(T)\rVert^2
\end{equation}
\begin{equation}
  \mathcal{E}_{\mathrm{norm}}(T) =
  \frac{\mathrm{Tr}(H_T^\top L H_T)}
       {\mathrm{Tr}((H_T - \bar{H}_T)^\top D (H_T - \bar{H}_T))}
\end{equation}
Lower $\mathcal{E}_{\mathrm{norm}}$
means adjacent nodes are closer together in activation space, a
representational signature that the model has internalized the graph's
adjacency structure beyond what token co-occurrence statistics alone would
produce. If the induction account is sufficient, we should see no coherent
graph structure emerge in the residual stream; if the belief account holds,
we expect the geometry to progressively mirror the true topology as $T$
increases past $N^*$.

\subsection{Residual-Stream Geometry Results}
The behavioral results suggest the LLM maintains a complexity-sensitive
structural bias. But do those behavioral signatures have a correlate inside the
model? Near $N^*$ ($T = 200$), the PC1/PC2 plane shows only partial ring
structure; by $T = 1400$ the ring topology is clearly recoverable in the
low-dimensional geometry, and $\mathcal{E}_{\mathrm{norm}}$ shifts from
$0.785$ at short context to $0.828 \pm 0.076$ at $T = 1400$. The internal
geometry and the behavioral phase transition move together.

\begin{figure}[t]
    \centering
    \includegraphics[width=0.9\columnwidth]
    {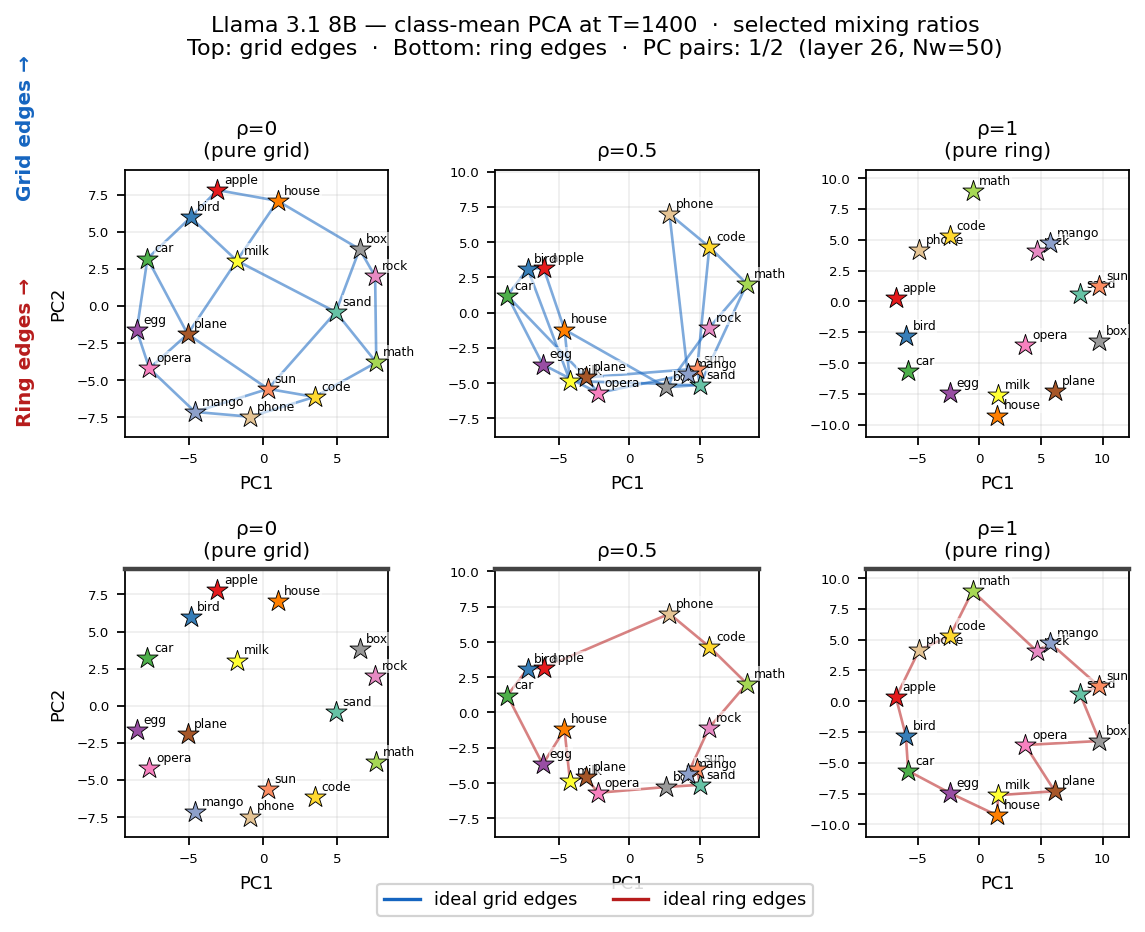}
    \caption{Snapshot of PCA embeddings for all tokens across value of mixture ratio $\rho$ (columns). 
    \emph{Top:} Blue edges of grid overlaid when grid has non-zero mixture weight.
    \emph{Bottom:} Same plots with red edges of ring overlaid.
    Full ladder of $\rho$ values in Appendix \ref{app:representation}
    }
    \label{figure-1}
\end{figure}

The stronger test comes from the competing-structures regime.
Figure \ref{figure-1} and Appendix~\ref{app:representation} shows class-mean PCA at $T = 1400$ across
the full $\rho$-ladder in a secondary closed-vocabulary experiment where grid
and ring share the same 16-token vocabulary. At
$\rho = 0.5$, both topologies are simultaneously encoded in orthogonal
subspaces. An induction circuit
accumulating local transition statistics would produce a single blended
representation, a mixture of grid and ring co-occurrences, not two separable
global structures in orthogonal subspaces.

\section{Experiment 3}
\subsection{Does Graph-Family Information Causally Control Next-Token Predictions?}
\label{sec:causal-methods}

The behavioral and PCA analyses are correlational: they show that outputs and
representations are consistent with latent structure inference, but not that the
relevant residual-stream information is used by the final prediction. We
therefore run two causal interventions, following the activation-patching and
activation-steering logic used in mechanistic interpretability and
representation engineering~\citep{meng2022locating,turner2023activation,zou2023representation}.

For clean graph $G_c$, corrupt graph $G_r$, and final token $x_t$, we score
each logit vector with a graph-family contrast
{\small
\begin{equation}
  \Delta(x_t) =
  \frac{1}{|\mathcal{N}_{G_c}(x_t)|}
  \sum_{w \in \mathcal{N}_{G_c}(x_t)} z_w
  -
  \frac{1}{|\mathcal{N}_{G_r}(x_t)|}
  \sum_{w \in \mathcal{N}_{G_r}(x_t)} z_w
  \label{eq:graph-logit-contrast},
\end{equation}
}
where $\mathcal{N}_{G}(x_t)$ denotes the set of valid next-node neighbors of $x_t$ under graph $G$, and $z_w$ is the next-token logit for word $w$.
We generate matched clean/corrupt prompt pairs from the grid and ring that end at the same current node. For activation patching, we cache clean residual activations and rerun
the corrupt prompt while replacing the final-position \texttt{hook\_resid\_post}
activation after block $\ell$. The normalized patch effect is
\begin{equation}
  E_{\mathrm{patch}}(\ell) =
  \frac{\Delta_{\mathrm{patch}}(\ell)-\Delta_{\mathrm{corrupt}}}
       {\Delta_{\mathrm{clean}}-\Delta_{\mathrm{corrupt}}}.
\end{equation}
Thus $0$ means no movement from the corrupt graph preference, and $1$ means the
patched corrupt run recovers the clean run's graph preference.

For steering, we compute a layer-wise graph-difference vector on disjoint
training contexts,
\begin{equation}
  v_\ell =
  \mathbb{E}[h_\ell(x_t)\mid G_c=\mathrm{grid}]
  -
  \mathbb{E}[h_\ell(x_t)\mid G_r=\mathrm{ring}],
\end{equation}
and add $\alpha v_\ell$ to final-position residual activations in held-out
ring contexts. We compare the real vector to two controls: a Gaussian random
vector matched to the real vector's norm and a shuffled-label vector computed
after permuting graph labels in the steering-vector training set. Full
protocol details, including the seen/held-out edge split used to test
transition-cache explanations, are in Appendix~\ref{app:causal-interventions}.

\subsection{Causal Activation Intervention Results}

Using the clean/corrupt prompt-pair protocol from Section~\ref{sec:causal-methods},
we intervene on final-position residual-stream activations and score the resulting
next-token distribution with the graph-family logit contrast in
Eq.~\ref{eq:graph-logit-contrast}.

We outline each finding briefly below; full details are in Appendix~\ref{app:causal-interventions}. At context length $T=1400$, final-token residual patching rises rapidly across the selected layer sweep, establishing that late residual-stream states causally control the graph-neighbor logit contrast rather than merely encoding decodable graph information. To test whether this effect reflects replay of locally observed transitions alone, we split clean graph neighbors into edges observed in the corrupt context and held-out edges. While the held-out effect is delayed, it crosses zero by layer 26 and reaches $2.0$ at layer 30, showing that patched activations boost graph-consistent predictions even for edges never observed in the corrupt prompt.

Steering provides a complementary lower-bandwidth causal intervention. Adding the grid-minus-ring direction to held-out ring contexts recovers $0.449 \pm 0.004$ of the clean-corrupt graph contrast at $\alpha=5$, while negative $\alpha$ reverses the effect and both random norm-matched and shuffled-label controls remain near zero. The effect also strengthens with layer. Because steering uses a single global direction rather than pair-specific activation replacement, it does not fully reproduce patching; Appendix~\ref{app:causal-interventions} shows that held-out edge-specific logits remain substantially harder to steer. Together, these interventions support the existence of a manipulable graph-family representation that contributes causally to next-token prediction.

\section{Discussion \& Conclusion}


The behavioral, representational, and causal evidence converge on the same
picture: Llama-3.1-8B is not well described as either a pure Bayesian structure
learner or a pure induction-cache machine. This does not rule out induction circuits. In fact, the delayed held-out edge
effect in patching and the incomplete held-out recovery under steering both
suggest that local transition evidence remains important. The more plausible
account is a dual-mechanism one: induction-like caches and latent-structure
representations operate together, with the residual stream integrating both
sources of evidence before the final prediction. Immediate next steps are to
measure subspace alignment angles against graph Laplacian eigenvectors across
$\rho$, run head-level ablations at the patching-identified layers, and scale
the same causal protocol to larger Llama models to test whether the recovered
complexity penalty $\hat\lambda$ grows with model capacity.

\newpage
\bibliography{refs}
\bibliographystyle{icml2026}

\newpage
\appendix
\clearpage
\onecolumn

\section{Behavioral Model Details}

\subsection{Baseline Derivation}
\label{app:baseline}

Let $S \in \{0, 1\}$ be a latent binary variable: $S=1$ indicates the LLM has
adopted the in-context graph structure; $S=0$ indicates reliance on pretrained
associations. The log-prior over $S$ is $b \in \mathbb{R}$, with $b < 0$
encoding initial skepticism toward the arbitrary in-context graph. As the LLM
observes more walk steps, evidence accumulates sub-linearly:
\begin{equation}
  \log\frac{p(\mathbf{x}\mid S=1)}{p(\mathbf{x}\mid S=0)} \approx \gamma N^{1-\alpha}
\end{equation}
where $\gamma > 0$ controls evidence strength per token and $\alpha \in (0,1)$
captures diminishing returns from correlated walk steps. Combining prior and
likelihood via Bayes' rule gives the predicted neighbor-hit accuracy at context
length $N$:
\begin{equation}
  \hat{p}(N) = p_0 + (q - p_0)\,\sigma\!\left(b + \gamma N^{1-\alpha}\right)
\end{equation}
where $p_0$ is the pre-transition neighbor-hit rate (estimated empirically from
$N \leq 100$ tokens), $q \in (p_0, 1]$ is the graph-mode success rate, and
$\sigma$ is the sigmoid. The phase transition inflection point is
$N^* = (-b/\gamma)^{1/(1-\alpha)}$, corresponding to the context length at
which log-odds cross zero and the LLM tips from skepticism to belief.

Parameters $\boldsymbol\theta = (b, \gamma, \alpha, q)$ are fit by minimizing
MSE between $\hat{p}(N)$ and observed accuracy curves, equivalent to MLE under
an additive Gaussian noise model on observed accuracies. We use L-BFGS-B with
16 random restarts and box constraints $b \in [-30, 30]$, $\gamma \in
[10^{-6}, 50]$, $\alpha \in [0, 0.99]$, $q \in (p_0, 1]$. The bounds enforce
domain constraints directly; the lowest-loss restart is kept and validation and
test MSE are reported afterward.

\subsection{Weighted Prior Model Estimation Details}
\label{app:upgrade}

The joint objective minimizes MSE over all $(\rho, k, N)$ triples:
\begin{equation}
  \hat{\boldsymbol\theta} = \arg\min_{\boldsymbol\theta}
  \sum_\rho \sum_k \sum_{N \in \mathcal{C}}
  \left[\hat{p}_{k,\text{obs}}(\rho, N) -
  \hat{p}_k(\rho, N;\boldsymbol\theta)\right]^2
\end{equation}
We use L-BFGS-B with 24 random restarts and box constraints $b_0 \in [-15,
15]$, $\lambda \in [-2, 2]$, $\gamma_k \in [10^{-6}, 50]$, $\alpha_k \in [0,
0.99]$, $q_k \in (p_{0,k}, 1]$. Note that $\lambda$ bounds include negative
values; $\hat\lambda < 0$ would mean the LLM prefers more complex graphs,
which would falsify the complexity-prior hypothesis.

Model selection between the per-graph (8 parameters) and mixture-bias (5
parameters) versions uses AIC and BIC under the Gaussian residual assumption:
\begin{equation}
  \text{AIC} = n\cdot(\log(2\pi\cdot\text{MSE})+1) + 2k, \quad
  \text{BIC} = n\cdot(\log(2\pi\cdot\text{MSE})+1) + k\log n
\end{equation}
where $n$ is the number of training observations and $k$ is the number of free
parameters. The pre-transition accuracy $p_{0,k}$ is estimated per graph by
averaging neighbor-hit accuracy over training walks at $N \leq 100$ tokens,
then pooled to a single $p_{0,\text{grid}}$ and $p_{0,\text{ring}}$ per
vocabulary condition for identifiability.

We note one limitation of the \texttt{overlap} fit: the optimizer saturated
the lower bound of the $b_0$ search range ($\hat{b}_0 = -15.00$), indicating
it wanted a more negative value than allowed. As a consequence,
$\hat\lambda_{\text{overlap}}$ and the implied $\hat{b}_{\text{grid}} -
\hat{b}_{\text{ring}}$ gap may be biased toward zero in this condition;
widening the bounds or reparameterizing $b_0$ is a straightforward follow-up.

\vfill
\pagebreak
\section{Representational Figures}
\label{app:representation}

\begin{figure}[H]
    \centering
    \includegraphics[width=\textwidth]{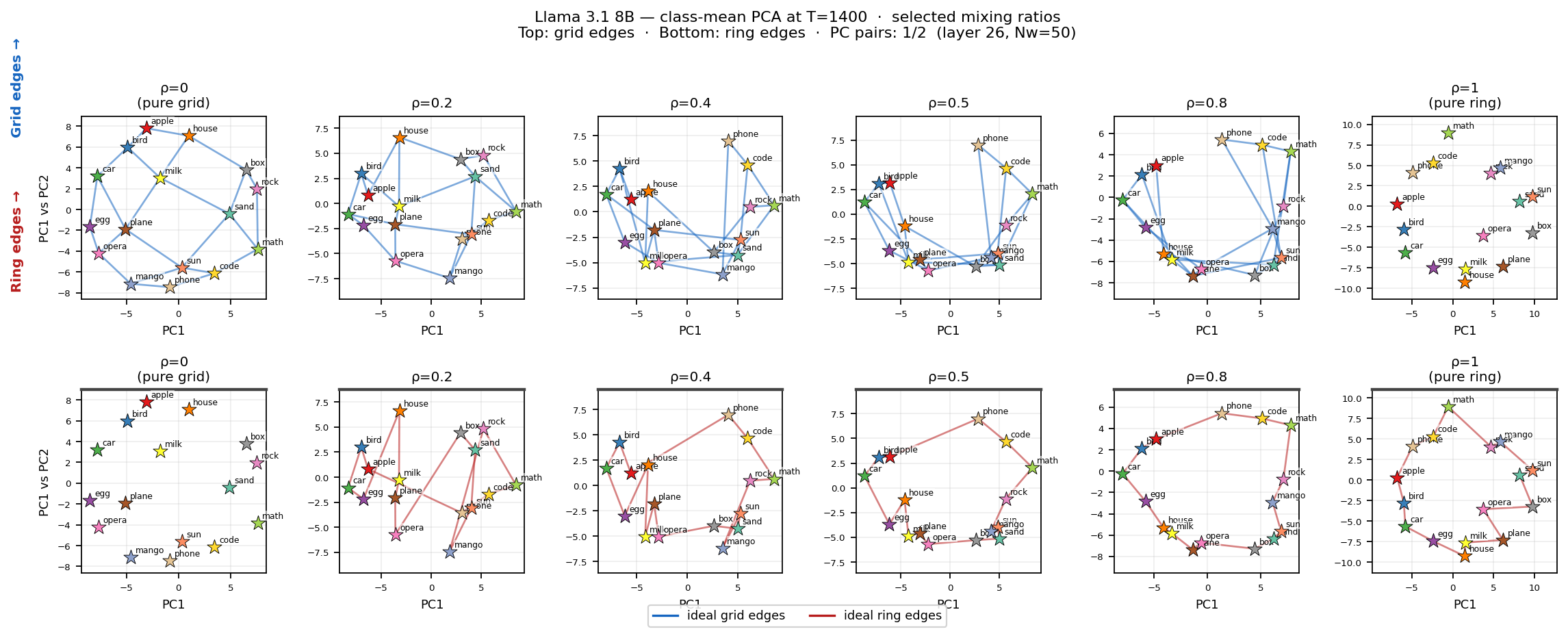}
    \caption{Full suite of PCA analysis on all mixture ratios $\rho$, with first row showing grid reconstruction edges in blue and second row with ring edges in red.}
    \label{fig:placeholder}
\end{figure}

\begin{figure}[H]
  \centering
  \includegraphics[width=0.86\linewidth]{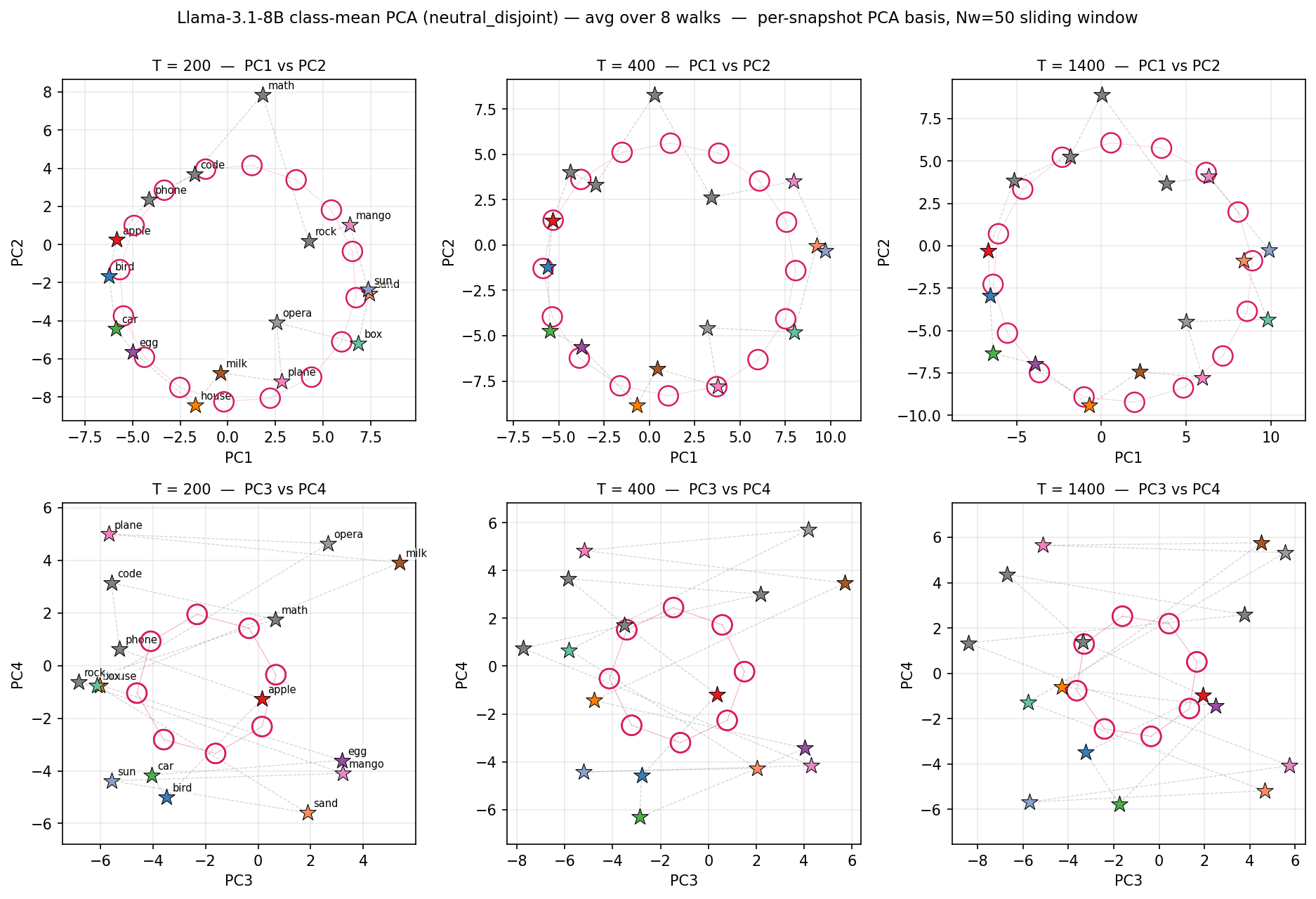}
  \caption{Representative class-mean PCA snapshots for the neutral disjoint
  vocabulary condition. These plots provide additional visual context for the
  layer-26 residual-stream geometry discussed in the main text.}
  \label{fig:pca-snapshots-appendix}
\end{figure}

\vfill
\pagebreak
\section{Causal Intervention Details}
\label{app:causal-interventions}

\subsection{Prompt Pair Construction and Metrics}

Clean and corrupt prompts are generated from different graph families but end
at the same current node. Because the graph hypotheses are undirected, we
sample a random walk ending at the desired final node by generating a valid walk
from that node and reversing it. The reversed walk has the same graph support,
and the model is always evaluated at the final position, predicting the next
graph word.

The primary score is the graph-family logit contrast in
Equation~\ref{eq:graph-logit-contrast}. For patching, normalized effect is
$(\Delta_{\mathrm{patch}}-\Delta_{\mathrm{corrupt}})/
(\Delta_{\mathrm{clean}}-\Delta_{\mathrm{corrupt}})$. For steering, normalized
effect is computed analogously, replacing $\Delta_{\mathrm{patch}}$ with the
steered metric and using the target and source prompt metrics as endpoints.
Rows with small denominators are marked unusable in the raw JSONL; none were
excluded in the reported patching runs.

\subsection{Seen and Held-Out Edge Split}

For each final token, we split graph neighbors according to whether the edge
incident to the final token was observed in the evaluation context. The
``seen'' set contains true graph neighbors of the final token whose edge
appeared in either direction. The ``held-out'' set contains true graph
neighbors whose edge did not appear. For clean/corrupt patching, this split is
computed using the corrupt context, so the diagnostic asks whether a clean
activation intervention helps graph-neighbor logits that the corrupt prompt did
not locally observe.

\subsection{Activation Patching}

The patching run reported in the paper uses Llama-3.1-8B, 200 grid/ring
clean-corrupt prompt pairs, context length $T=1400$, final-position
\texttt{hook\_resid\_post} activations, and the selected layer set
$\{14,15,16,20,24,26,28,30\}$.

\begin{figure}[H]
  \centering
  \begin{subfigure}[t]{0.48\linewidth}
    \centering
    \includegraphics[width=\linewidth]{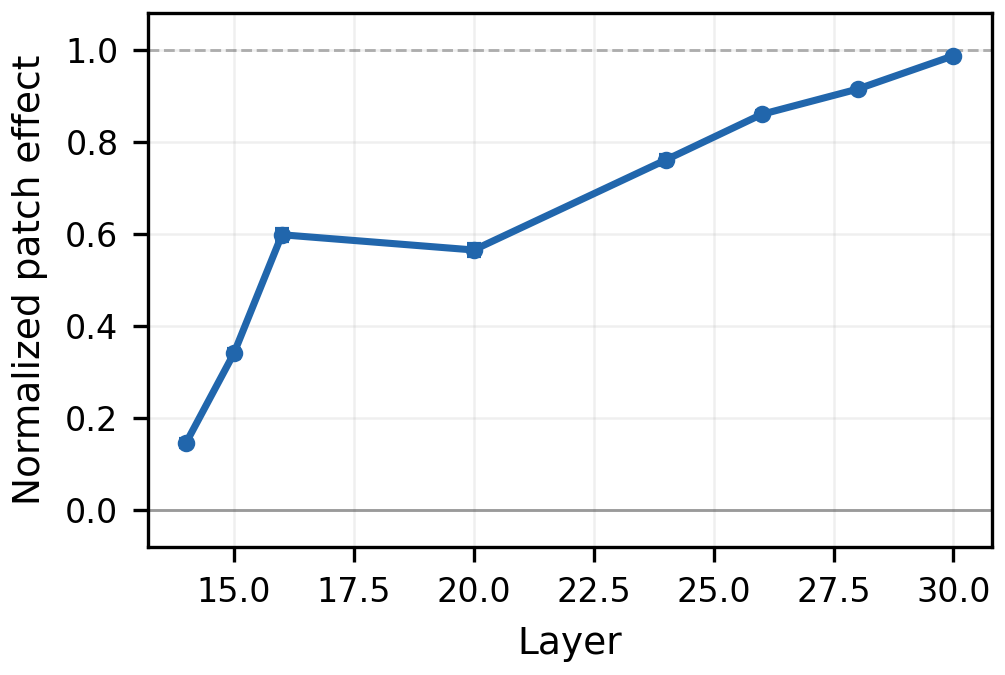}
    \caption{Final-token patching at $T=1400$.}
  \end{subfigure}
  \hfill
  \begin{subfigure}[t]{0.48\linewidth}
    \centering
    \includegraphics[width=\linewidth]{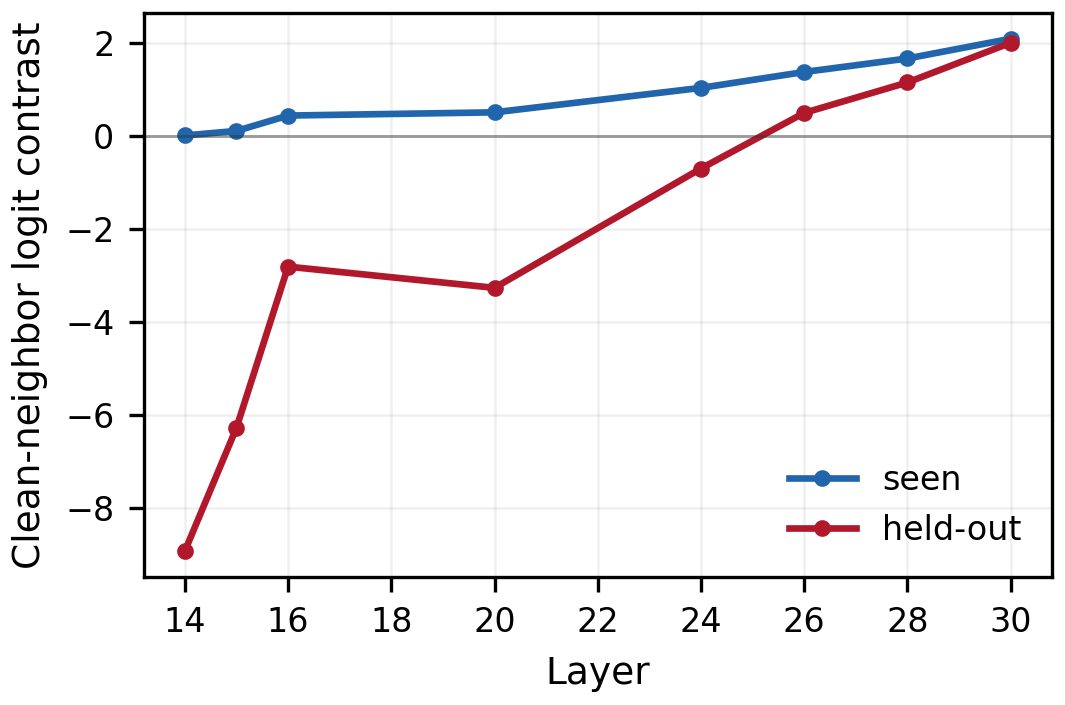}
    \caption{Seen/held-out split at $T=1400$.}
  \end{subfigure}
  \caption{Long-context activation-patching diagnostics. The selected-layer
  run shows late-layer recovery and held-out graph-neighbor logits becoming
  positive in late layers.}
  \label{fig:patching-long-appendix}
\end{figure}

\begin{table}[H]
\footnotesize
  \centering
  \caption{Selected activation-patching aggregates. Effects are mean $\pm$
  standard error over prompt pairs.}
  \label{tab:patching-aggregates}
  \begin{tabular}{ccccc}
    \toprule
    Context & Layer & $n$ & Normalized effect & Held-out contrast \\
    \midrule
    1400 & 16 & 200 & $0.598 \pm 0.013$ & $-2.814$ \\
    1400 & 26 & 200 & $0.860 \pm 0.008$ & $0.492$ \\
    1400 & 30 & 200 & $0.987 \pm 0.001$ & $2.000$ \\
    \bottomrule
  \end{tabular}
\end{table}

\subsection{Activation Steering}

The steering run uses disjoint train and evaluation contexts. We compute
grid-minus-ring vectors from 1000 training contexts per graph at $T=1400$,
then evaluate on 500 held-out grid/ring prompt pairs at layers 20--28 and
$\alpha \in \{-5,-2,-1,-0.5,0,0.5,1,2,5\}$. The raw JSONL contains duplicate
rows from a prior append; all reported steering aggregates and paper figures
deduplicate by the complete intervention key, keeping the first occurrence
(\texttt{pair\_id}, layer, $\alpha$, control, evaluation direction), yielding
the expected 243,000 unique interventions.

\begin{figure}[H]
  \centering
  \begin{subfigure}[t]{0.32\linewidth}
    \centering
    \includegraphics[width=\linewidth]{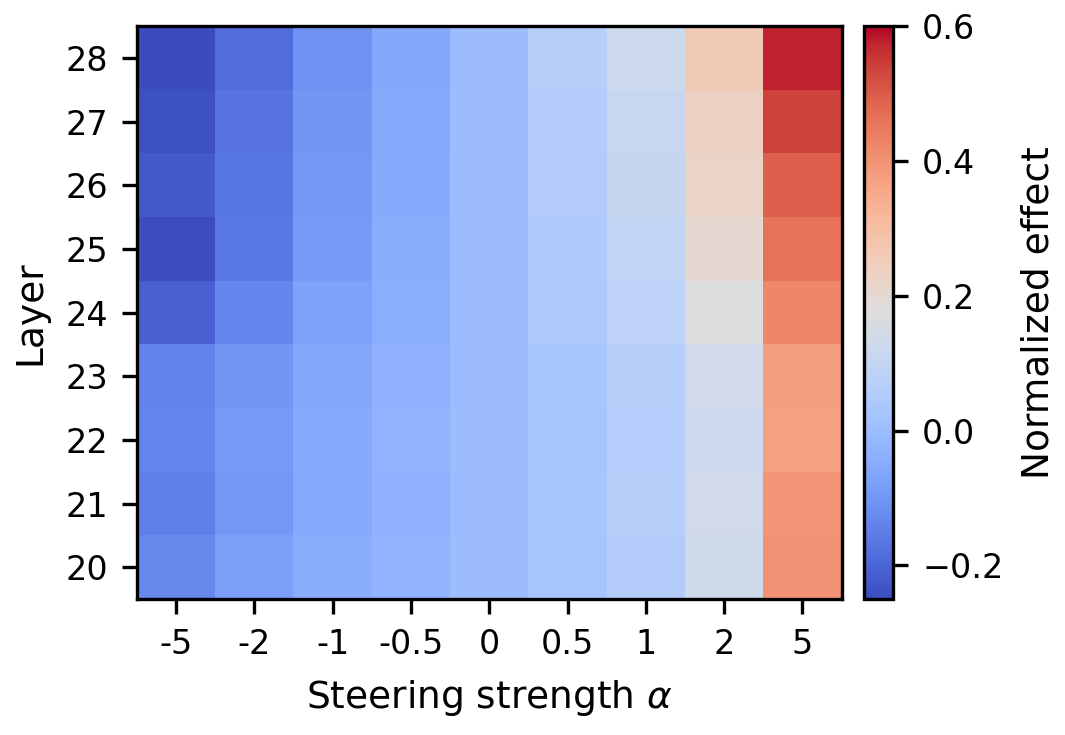}
    \caption{Target-to-source layer/alpha heatmap.}
  \end{subfigure}
  \hfill
  \begin{subfigure}[t]{0.32\linewidth}
    \centering
    \includegraphics[width=\linewidth]{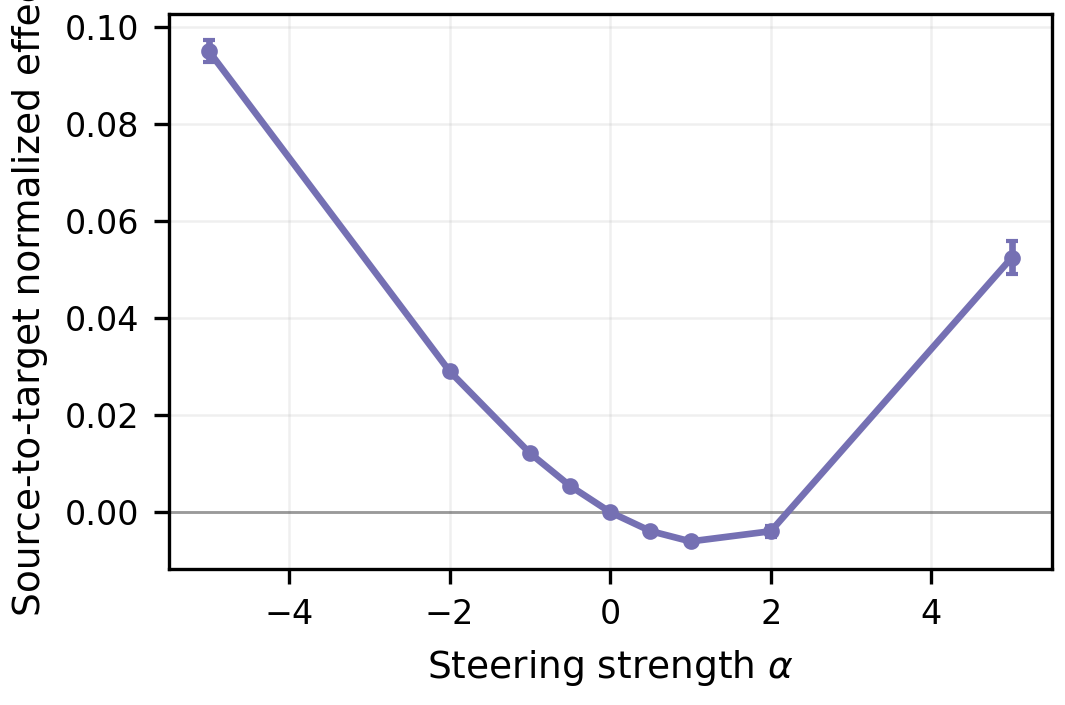}
    \caption{Source-to-target direction.}
  \end{subfigure}
  \hfill
  \begin{subfigure}[t]{0.32\linewidth}
    \centering
    \includegraphics[width=\linewidth]{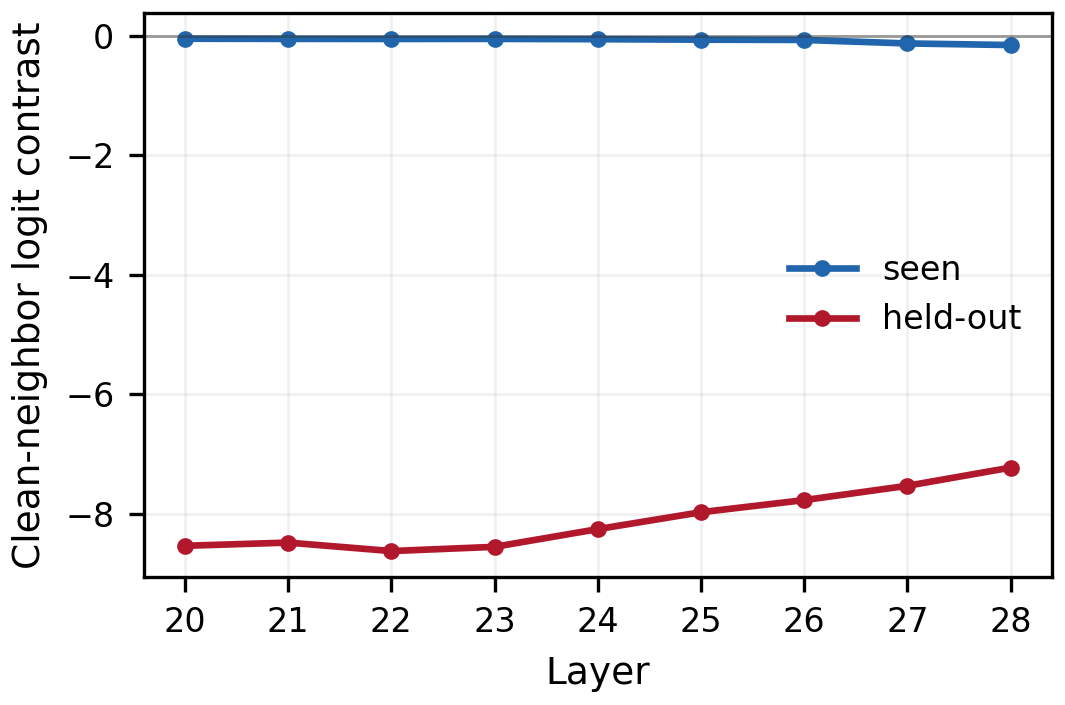}
    \caption{Positive-alpha seen/held-out split.}
  \end{subfigure}
  \caption{Additional steering diagnostics. Steering is strongest for
  target-to-source grid-minus-ring additions in late layers. The reverse
  direction is weaker, and held-out edge-specific logits remain difficult to
  move with a single global vector.}
  \label{fig:steering-appendix}
\end{figure}

\begin{table}[H]
  \centering
  \caption{Deduplicated target-to-source steering aggregates, averaged over
  layers 20--28 and 500 evaluation pairs.}
  \label{tab:steering-alpha}
  \begin{tabular}{cccc}
    \toprule
    $\alpha$ & Real vector & Random norm-matched & Shuffled labels \\
    \midrule
    $-5$ & $-0.192 \pm 0.003$ & $0.016 \pm 0.001$ & $0.010 \pm 0.000$ \\
    $0.5$ & $0.042 \pm 0.000$ & $0.000 \pm 0.000$ & $-0.001 \pm 0.000$ \\
    $1$ & $0.087 \pm 0.001$ & $0.001 \pm 0.000$ & $-0.002 \pm 0.000$ \\
    $2$ & $0.180 \pm 0.002$ & $0.003 \pm 0.000$ & $-0.003 \pm 0.000$ \\
    $5$ & $0.449 \pm 0.004$ & $0.019 \pm 0.001$ & $-0.006 \pm 0.000$ \\
    \bottomrule
  \end{tabular}
\end{table}

\paragraph{Interpretation.}
Activation patching is a high-bandwidth, pair-specific intervention: it
replaces the corrupt residual state with the clean residual state at one layer
and position. Steering is a low-bandwidth, population-level intervention: it
adds one global graph-difference vector to many contexts. The large gap between
near-complete patching recovery and partial steering recovery is therefore not
a contradiction. It suggests that the residual stream contains both a broad
graph-family direction and finer edge- or context-specific information that a
single steering vector does not capture.

\end{document}